\documentclass[12pt]{article}
\usepackage[utf8]{inputenc}
\usepackage{amsmath,amsfonts,amssymb}
\usepackage[margin=1.0in]{geometry}
\usepackage{graphicx}
\usepackage[font=small,labelfont=bf]{caption}
\usepackage[final]{pdfpages}
  \DeclareGraphicsExtensions{.png,.pdf}
\usepackage[section]{placeins}
\usepackage[symbol]{footmisc}
\usepackage{authblk}

\renewcommand{\thefootnote}{\fnsymbol{footnote}}

\newcommand\blfootnote[1]{%
  \begingroup
  \renewcommand\thefootnote{}\footnote{#1}%
  \addtocounter{footnote}{-1}%
  \endgroup
}

\usepackage{biblatex} 
\usepackage{hyperref}
\hypersetup{
    colorlinks=true,
    linkcolor=blue,
    filecolor=magenta,      
    urlcolor=cyan,
    pdftitle={Overleaf Example},
    pdfpagemode=FullScreen,
    }

\usepackage{listings}
\usepackage{color}

\definecolor{dkgreen}{rgb}{0,0.6,0}
\definecolor{gray}{rgb}{0.5,0.5,0.5}
\definecolor{mauve}{rgb}{0.58,0,0.82}

\title{TorchNTK: A Library for Calculation of Neural Tangent Kernels of PyTorch Models}
\author[1]{Andrew Engel}
\author[2]{Zhichao Wang}
\author[3]{Anand D. Sarwate}
\author[1]{Sutanay Choudhury}
\author[1]{Tony Chiang}

\small
\affil[1]{Pacific Northwest National Laboratory, Richland, WA, 99354, USA}
\affil[2]{University of California San Diego, La Jolla, CA 92093, USA}
\affil[3]{Rutgers University - New Brunswick}
\date{}

\begin{document}

\maketitle

\blfootnote{Corresponding Authors: Engel, A., andrew.engel@pnnl.gov and Chiang, T., tony.chiang@pnnl.gov}

\begin{center}
    \textbf{Abstract}
\end{center}

We introduce \textbf{torchNTK}, a python library to calculate the empirical neural tangent kernel (NTK) of neural network models in the PyTorch framework. We provide an efficient method to calculate the NTK of multilayer perceptrons. We compare the explicit differentiation implementation against autodifferentiation implementations, which have the benefit of extending the utility of the library to any architecture supported by PyTorch, such as convolutional networks. A feature of the library is that we expose the user to layerwise NTK components, and show that in some regimes a layerwise calculation is more memory efficient. We conduct preliminary experiments to demonstrate use cases for the software and probe the NTK. Our software can be installed from Github, \href{https://github.com/pnnl/torchntk}{ here.} 


\section{Introduction}
Artificial neural networks (ANNs) give unprecedented results in machine learning tasks \cite{AlexNet2012, resnet2015, InceptionResNetV4, ViTOG, EffNetV2OG} though they continue to be little understood. Theoretical studies focus on the equivalence between ANNs trained with (full-batch) gradient descent and kernel methods \cite{Jacot_NTK2018_Convergence, CNTK_Sanjeev, jaxgithub, Domingos21}. Specifically, it was shown that in the infinite-width limit neural networks trained by gradient descent are equivalent to a kernel machine using the NTK \cite{Jacot_NTK2018_Convergence}. It was then shown that the necessary condition for the equivalence of a neural network model to kernel regression was being in a so called 'lazy-training' regime where the weights remain approximately static \cite{LazyTraining}. This is equivalent to using a linearization of the neural network model about its initial parameterization \cite{LazyTraining, WideNNevolveaslinear}. Critics argue that being within this 'lazy-training' regime gives results in-consistent with phenomena that we observe, such as feature learning \cite{TensorPrograms4}, and that using the linearization of a model around its initialization can degrade performance significantly \cite{LazyTraining}. Following this, many teams evaluated the difference and similarities between kernel regression using NTKs and neural networks across a variety of tasks \cite{CNTK_Sanjeev, FinitevsEmpirical2020, HarnessingThePowerArora2020}, but generally found that the performance differences depended on task and architecture \cite{DisentanglingFeatureLazy}. 

Our team became interested in calculating these kernels for neural network models of used in practice. A python library has been developed to dramatically increase the ease of calculating both finite and infinite width networks' tangent kernels called neural-tangents \cite{neural_tangents}, but this library has a few limitations. First, it is built upon the jax framework \cite{jaxgithub}. Jax is not currently natively supported on windows machines, and is still in pre-release. Jax has not yet developed as large a user-base as other popular frameworks like TensorFlow \cite{tensorflow} or PyTorch \cite{PyTorch} (see for example: \cite{kaggle_2021}), and until a time where it does so, using neural-tangents will require users to overcome the hurdle of mastering another deep learning framework. A different code-base that uses cupy \cite{cupy} was built to efficiently calculate the infinite convolutional NTK \cite{CNTK_Sanjeev}, but not empirical NTKs. Therefore, a niche role in the community was present to provide software to calculate the tangent-kernel within a framework that is widely used in the field, PyTorch.



 We present torchNTK, a python library built on the PyTorch framework that calculates the empirical NTK using PyTorch autograd, and explicitly for multilayer perceptron networks. We developed torchNTK to achieve our goals of studying the time-evolution of the NTK for models that practical users of AI systems interface with. In the remainder of this paper, we describe the explicit algorithms used to calculate the NTK, we benchmark our software to compare implementations, detail an initial experiment to demonstrate our software, and discuss plans for improving our software. 
 
 
 
 \section{The Additive Components of the Neural Tangent Kernel} \label{AdditiveComponents}
 
 Study of the NTK of finite-networks of large size trained on large number of datapoints has been difficult due to the sizes of the matrices involved. If a neural network parameterized by $\theta$ and acts on a dataset X is denoted by $f_{\theta}(X)$ then, the NTK matrix is the Gram matrix of the Jacobian of the network \cite{Jacot_NTK2018_Convergence} as follows:
 \begin{equation}
     J = (\nabla_{\theta} f_{\theta}(X)),\quad K^{NTK} =  J^{\top}J.
 \end{equation}

 
 For a dataset of size $N$ and a network of size $P$, the Jacobian is a size $P\times N$ matrix. Considering that deep learning is applied to problems that generally have many datapoints and models have increased in size over time, the full Jacobian matrix is often too costly to hold in memory on consumer workstations. As a concrete example, for a dataset of size 60,000 and a network with 100,000 parameters represented in fp32, the Jacobian is a 24 gigabyte matrix, which is larger than the total available VRAM on most GPUs. Note, that these sizes are typical for common toy problems like digit classification but that modern architectures might have 1e1 - 1e6 times the number of parameters \cite{GPT3, resnet2015}. While the Jacobian is large due to the number of parameters, the NTK is size $N\times N$, and for modestly sized datasets the NTK is more realistic to expect to hold in memory.
 
 In the over-parameterized regime, we can lower peak memory requirements by transforming the problem from holding a $P\times N$ matrix into holding many $N\times N$ matrices using a layerwise approach. These additive components have already been pointed out directly in works that derive algorithms for the calculation of the NTK \cite{Zhichaos_paper, CNTK_Sanjeev, Lee2019_empricalNTK}, and can be most explicitly represented as a sum over the layers $L$:
 \begin{equation}
     K^{\mathrm{NTK}} = \sum_{l=0}^L (\nabla_{\theta^l} f_{\theta)}(X)) (\nabla_{\theta^l} f_{\theta}(X))^{\top}.
 \end{equation}

We hypothesize that the additive components representing the layers contain more specific information about the operations they represent and may be getting 'lost' in the full NTK, though we leave demonstrating that to future work. For that reason, these components are worthy of additional study, and in fact, there has been recent work on a spectral analysis of these layerwise kernels \cite{LayerwiseSpectral}.

We end this section by pointing out that in contemporary works on the weight matrices \cite{MartinMahoneyNature}, the Hessian \cite{LayerwiseHessian,Layerwise_Hessian}, and Fisher information matrices \cite{Karakida3_crossentropy}, have all examined a 'layerwise' approach. These additive components of the NTK can be thought of as a natural extension to compliment these modes of inquiry. 

\section{Algorithmic Details}
TorchNTK is an accumulation of different methods to calculate the NTK, which can be broadly classified as either autograd or explicit differentiation. While autograd methods can handle any model, the explicit differentiation technique was benchmarked to be much faster on the MLP architectures that it is limited to. 
In the following sections we derive the formula for MLP architectures used to recursively calculate the NTK:

\subsection{Derivation of the NTK: MLP without bias}

Consider the following to represent a neural network with parameters $\theta$ and $X \in \mathbf{R}^{d_0,n}$ where $d_0$ is the dimensionality of the input vector, or equivalently is the width of the input layer to our neural network. Let us first consider neural networks composed of a series of matrix multiplications, interrupted by non-linear activation functions. These networks are referred to as multilayer perceptrons. Below, we also use the convention that $X_{l}$ is the output of layer $l$, that $\sigma$ is some activation function, and that $W$ is the weight matrix. 

$$f_{\theta}(X) = \mathbf{w}^{\top} \frac{1}{\sqrt{d_L}}\sigma(W^{\top}_L \frac{1}{\sqrt{d_{L-1}}} \sigma (... \frac{1}{\sqrt{d_2}} \sigma(W^{\top}_2 \frac{1}{\sqrt{d_1}} \sigma(W^{\top}_1 \mathbf{X}))))$$ 

Where we have adopted the practice of dividing by the square root of the width of each layer which is necessary to place ourselves in the kernel parameterization.  

Given that $S_l$ is a matrix whose $\alpha$th column is:
$$s_{\alpha}^l = D_{\alpha}^l \frac{W_{l+1}^\top}{\sqrt{d_l}}  D_{\alpha}^{l+1} \frac{W_{l+2}^\top}{\sqrt{d_{l+1}}} D_{\alpha}^{l+2} ...  \frac{W_{L}^\top}{\sqrt{d_{L-1}}}  D_{\alpha}^{L} \frac{\mathbf{w}}{\sqrt{d_L}},$$
where D is defined by
$$ D_{\alpha}^{k} \equiv \mathrm{diag}( \sigma^\prime(W_{k} x_{\alpha}^{k-1} )  ) $$
Then one can show (see Appendix G.2 of \cite{Zhichaos_paper}):

$$K^{NTK} = X^\top_L X_L + \sum_{l=1}^L (S_l^\top S_l) \odot (X_{l-1}^\top X_{l-1})$$

The NTK is therefore a sum over components, each themselves being the product of a co-variance matrix of features preceding the layer and a term related to the propagation of gradients inside the network.

\subsection{Derivation: MLP with bias}

We extend these results to include a bias vector:

$$f_{\theta}(x) = \mathbf{w}^{\top} \frac{1}{\sqrt{d_L}}\sigma(W^{\top}_L \frac{1}{\sqrt{d_{L-1}}} \sigma (... \frac{1}{\sqrt{d_2}} \sigma(W^{\top}_2 \frac{1}{\sqrt{d_1}} \sigma(W^{\top}_1 \mathbf{x} + B_1) +B_2) ... +B_{L-1}) +B_L) + \mathbf{B}$$ 

We need to update our terms as well, so that each layers output is now:

$$X_{l} = \frac{1}{\sqrt{d_l}} \sigma(W^{\top}_{l} X_{l-1} + B_{l})$$

And update our definition of $D_l$:

$$D_{l} = \mathrm{diag}(\sigma^{\prime}(W_l^{\top} X_{l-1} + B_{l}))$$

We can now derive the equation for the bias vectors' contribution to the NTK for the bias of any layer, $B_{l}$. Taking the series of gradients of each weight bearing tensor in the operation reveal that the bias vectors also contribute components equal to the matrix S described above times the element wise product with a matrix of all ones, $J_{n}$, which we describe below.

$$ \frac{\partial f_{\theta}(x)}{\partial B_{l}} = \mathbf{w}^{\top} \frac{1}{\sqrt{d_{L}}} \sigma^{\prime}(W_{L}^\top X_{L-1} + B_{L}) W_{L-1}^{\top} ... \sigma^{\prime}(W_{l}^\top X_{l-1} + B_{l}) \frac{\partial B_{l}}{\partial B_{l}} $$

Substituting in the definition for S taken above (with our new definition of D):

$$ \frac{\partial f_{\theta}(x)}{\partial B_{l}} = S_l \frac{\partial B_{l}}{\partial B_{l}}$$

Considering that $B_{l}$ is a matrix $\in \mathbb{R}^{d_{l}, n}$, and that $S_{l}$ is the same for every element inside that matrix, we can represent the computation for the entire matrix of bias parameters as an element wise product with the matrix of ones, $J$. This matrix is also $\in \mathbb{R}^{d_{l}, n}$. We will notate the first dimension as a subscript and leave the second dimension understood. Thus, $J_{d_{l}}$ is the matrix of all ones with shape $\mathbb{R}^{d_{l}, n}$. This allows us to write:

$$ \frac{\partial f_{\theta}(x)}{\partial B_{l}} = S_l \odot J_{d_{l}}$$

The NTK component from the bias at layer $l$ is therefore:

$$K^{NTK}_{B_l} = (S_l \odot J_{d_{l}})^{\top} (S_l \odot J_{d_{l}})$$ 

The total NTK can therefore be expressed as:

$$ K^{NTK} = X^\top_L X_L + J_{d_L}^\top J_{d_L}  + \sum_{l=1}^L (S_l^\top S_l) \odot (X_{l-1}^\top X_{l-1}) + \sum_{l=1}^L (S_l \odot J_{d_l} )^\top (S_l \odot J_{d_l} ) $$

\subsection{Autograd Algorithms}
In addition to the algorithm described above that explicitly calculates the NTK for MLP architectures, there are additional algorithms included in torchNTK that calculate the NTK using autograd methods. Autograd methods, while slower than our explicit NTK calculation, extend to other PyTorch architectures and largely work 'out-of-the-box' reducing the user's margin of error:

\begin{itemize}
    \item [1. ] The first alternative makes a call to \textit{torch.autograd.functional.jacobian} across the dataset, stacks the resulting list of tensors from each datapoint, then simply constructs the NTK as: 
    $$K^{NTK} = \nabla f(x,\theta)^{\top} \nabla f(x,\theta)$$ 
    \item[2. ] A second alternative calls autograd on the model iteratively across each layer for each datapoint, and was adapted from the work of \cite{TENAS}
    \item[3. ] A third alternative computes each row of the Jacobian vector product for each operation, then outputs each operation to a dictionary. This represents our 'layerwise' autograd method
    \item[4. ] With PyTorch 1.11, a new \textit{torch.vmap} function was created to parallelize computations across the batch dimension. One specific use case is to speed up the computation of the Jacobian. This can also be applied to speed up the computation of the layerwise autograd method and we have included it as a piece of experimental software.
\end{itemize}

\section{Software Performance}
In this section we detail the performance differences and trade-offs between the various algorithms for two classes of models: a MLP and a CNN, at two different widths. All algorithms were bench marked for their time to completion and maximum GPU memory allocations for calculating the final NTK of the same model for the same data on the same hardware inside an IPython kernel. We tested the algorithms on a local computer cluster equipped with an A-100 DGX node which we queried for 4 cpu cores and just one A-100 GPU. All algorithms were tested using GPU tensors and models, except the \textbf{Full Jacobian} implementation, which is CPU only.

For each of the 'MLP' benchmarks, we created a neural network represented by a Module object with 4 fully-connected layers. Each hidden layer had a width of 100 neurons. The input data was a vector of length 100 drawn from the standard normal distribution. The network terminated into a single neuron. Each hidden layer used the tanh activation function, while the output neuron was not routed through an activation function before calculating the NTK. As is common in the NTK literature, we used the NTK parameterization by dividing each layer's output by the square root of the width of that layer. The weights between each benchmark were kept the same using a random number generator seed, and were themselves drawn from the standard normal distribution.

As a check of our claim that we expect layerwise computations to be more memory efficient in the deep and narrow regime, we also calculated a memory benchmark for a high parameter MLP, called 'MLP-h'. This model has 8 layers, each hidden layer with a width of 1000, an input shape of 1000, tanh activation functions, terminates into a single neuron output. All memory benchmarks were calculated with the maximum allocated memory on the GPU, (\textit{torch.cuda.max\_memory\_allocated}()).




\begin{table}[!htbp]
\begin{center}
\title{MLP Benchmark Time}
\resizebox{\textwidth}{!}{%
\begin{tabular}{| c | c | c | c | c | c |}
\hline
\textbf{N Datapoints} & \multicolumn{5}{ c |}{\textbf{Time [sec]}}  \\ 
\cline{2-6}
& \textbf{Full Jacobian} & \textbf{Autograd} & \textbf{L. Autograd} & \textbf{L. Autograd w vmap} & \textbf{Explicit Differentiation} \\
\hline
10 & 3.3e-3 & 5.95e-3 & 18.2e-3 & 3.97e-3 & 1.01e-3 \\ \hline
100 & 39e-3 & 47.8e-3 & 174e-3 & 10.3e-3 & 0.98e-3 \\ \hline
1000 & 3.71 & 467e-3 & 1.97 & 90.4e-3 & 1.06e-3 \\ \hline
10000 & 359 & 4.77 & 19.8 & 869e-3 & 9.03e-3 \\ \hline
30000 & OOM & 15.7 & 63.0 & 2.89 & 74.7e-3 \\ \hline
40000 & OOM & 21.7 & 84.1 & 4.06 & OOM \\ \hline
\end{tabular}}

\caption{For a variety of number of datapoints, we ran our empirical NTK calculation algorithms on the same MLP model and calculated the time until completion using the IPython magic function \% timeit and report the mean time here. For function calls longer than 10 seconds a single function call's time is reported, calculated using the difference in system clock time. When the algorithm we called failed due to running out of system memory, we report 'OOM' instead. Generally we see that among autograd methods, the algorithms making use of DataLoader objects to feed the calculation (Autograd and L. Autograd with vmap) are faster, with torch.vmap enabling greater parallelization. With any appreciable amount of data, the method that calculates the full Jacobian becomes appreciably slower, so much so, that for the next set of benchmarks on a larger model we exclude the algorithm from analysis (Tables \ref{T:MLP-h Benchmark Time} and \ref{T:MLP-h Benchmark Memory}). Explicit Differentiation is faster than the alternatives, owing this to the engineering made to parallelize the computation across the entire dataset. With greater parallelization generally comes higher memory costs, which is explored further in Table \ref{T:MLP Benchmark Memory}.}
\label{T:MLP Benchmark Time}
\end{center}
\end{table}

\begin{table}[!htbp]
\begin{center}
\title{MLP Benchmark Memory}
\resizebox{\textwidth}{!}{%
\begin{tabular}{| c | c | c | c | c |}
\hline
\textbf{N Datapoints} & \multicolumn{4}{ c |}{\textbf{Memory [Mb]}}  \\ 
\cline{2-5}
& \textbf{Autograd} & \textbf{L. Autograd} & \textbf{L. Autograd w vmap} & \textbf{Explicit Differentiation} \\
\hline
10  & 2.82 & 1.61 & 1.24 & 0.45 \\ \hline
100  & 25.39 & 12.89 & 13.31 & 1.3  \\ \hline
1000  & 242 & 132 & 94.27 & 32.15 \\ \hline
10000  & 2416 & 2042 & 2471 & 1659  \\ \hline
30000  & 7250 & 14514 & 21809 & 14572  \\ \hline
40000  & 11233 & 25751 & 25731 & OOM \\ \hline
\end{tabular}}
\caption{Each algorithm (Except for Full Jacobian which was implemented only on the CPU, so our benchmark technique was not recorded) was run once on the GPU and the peak memory allocated on the GPU device was recorded. If the device ran out of memory, then 'OOM' was recorded instead. There are a few factors that are coupled that determine the memory usage: including whether the computation is batched over the number of data points, whether the computation parallelizes over the number of datapoints, whether the computation is layerwise or not, the architecture, and the number of datapoints. Due to this complexity, we suggest trying each method on a subset of the data first to search for a suitable choice.}
\label{T:MLP Benchmark Memory}
\end{center}
\end{table}

\begin{table}[!htbp]
\begin{center}
\title{MLP-h Benchmark Time}
\resizebox{\textwidth}{!}{%
\begin{tabular}{| c | c | c | c | c |}
\hline
\textbf{N Datapoints} & \multicolumn{4}{ c |}{\textbf{Time [sec]}}  \\ 
\cline{2-5}
& \textbf{Autograd} & \textbf{L. Autograd} & \textbf{L. Autograd w vmap} & \textbf{Explicit Differentiation} \\
\hline
10 & 25.3e-3 & 70.5e-3 & 30.8e-3 & 2.43e-3 \\ \hline
100 & 234e-3 & 690e-3 & 184e-3 & 2.36e-3  \\ \hline
500 & 1.31 & 3.38 & 1.3 & 2.08e-3 \\ \hline
1000 & OOM & 6.89 & 2.29 & 2.13e-3  \\ \hline
10000 &  OOM & OOM & OOM & 44.3e-3  \\ \hline
20000 &  OOM & OOM & OOM & 161e-3 \\ \hline
\end{tabular}}
\caption{Given our belief that a layerwise computation will be memory efficient in the deeper and narrower regime, we re-ran our benchmarks on a deeper model for comparison. Each benchmark was calculated using the IPython magic timieit function, with mean times reported here. If the calculation took longer than 10 seconds, a single calculation was used with the time computed from the difference in system clock times.  Because MLP-h has many more parameters, the size of the Jacobians needed to calculate the NTK more quickly saturate the available GPU memory. Whenever the calculation used all the available GPU memory we replace the value with 'OOM'. We see here that in the deeper regime, our layerwise approaches extend the calculation as we expected}
\label{T:MLP-h Benchmark Time}
\end{center}
\end{table}

\begin{table}[!htbp]
\begin{center}
\title{MLP-h Benchmark Memory}
\resizebox{\textwidth}{!}{%
\begin{tabular}{| c | c | c | c | c |}
\hline
\textbf{N Datapoints} & \multicolumn{4}{ c |}{\textbf{Memory [Mb]}}  \\ 
\cline{2-5}
& \textbf{Autograd} & \textbf{L. Autograd} & \textbf{L. Autograd w vmap} & \textbf{Explicit Differentiation} \\
\hline
10 & 717 & 277 & 238 & 158 \\ \hline
100 & 5695 & 1314 & 903 & 101  \\ \hline
500 & 28101 & 6227 & 4160 & 175 \\ \hline
1000 & OOM & 12378 & 8238 & 284  \\ \hline
10000 &  OOM & OOM & OOM & 4983  \\ \hline
20000 &  OOM & OOM & OOM & 17899 \\ \hline
\end{tabular}}
\caption{We also include memory benchmarks from each algorithm with increasing number of datapoints, where we can see that in the deeper and narrower regime this model probes, and especially with many datapoints, layerwise based calculations use less peak GPU memory than non-layerwise approaches. Our Explicit Differentiation model extended the calculation of the finite NTK to 10000 datapoints under 5GB of peak allocated memory, demonstrating that our code extends the study of the finite NTK with many datapoints to modest workstations. There exist many consumer level GPU choices with more than 5GB of memory.}
\label{T:MLP-h Benchmark Memory}
\end{center}
\end{table}



The results of these tables demonstrate that for very deep MLP networks, our explicit differentiation technique is more memory efficient and many times more time efficient than autograd methods. In addition, we show there is a regime of model architectures where layerwise computations are more memory efficient than full Jacobian computation with Autograd. Because the 'best choice' of algorithm depends on the specific goal of the researcher, we emphasize that individual researchers should benchmark their own architectures on their own systems to make an informed decision about which NTK algorithm suits them. Another key takeaway is that more effort should be placed into developing and investigating highly optimized explicit differentiation techniques for other model architectures. It is clear that tremendous benefits exist in doing so: our MLP-h model benchmark shows a speed up of over 1000x compared to the nearest autograd technique on 1000 datapoints. While tedious, in scenarios where limited hardware is available or has a high cost, or where many of these NTKs will need to be calculated (for instance, see our experiments in S\ref{VisNTKoverTraining} below) the benefits of explicit differentiation can outweigh the up-front development costs. Finally, other neural tangent libraries may benefit in reducing peak memory use with a layerwise approach.


\section{Experiments and Use Cases}

\subsection{Fisher Information Matrix}
As pointed out in contemporary work on the Fisher Information Matrix, the NTK shares its non-zero eigenvalues with its dual matrix \cite{Karakida2_mse,Karakida3_crossentropy}: 

$$F = \left(\frac{\partial f_{\theta}(x)}{\partial \theta}\right)\left( \frac{\partial f_{\theta}(x)}{\partial \theta}\right)^\top$$

The Fisher information matrix (FIM) is of interest because at convergence with training loss zero the Hessian of the mean squared loss function is equal to the FIM. In the following equation t indexes the datapoint inside a dataset of size T, see equation 9 of \cite{Karakida2_mse}.

$$H = F - \frac{1}{T} \sum_{t}^{T} (y(x_t) - f(x_t)) \nabla_{\theta} \nabla_{\theta} f(x_t)$$

 This makes the FIM useful in study the geometry of the loss landscape. Authors have suggested studying the eigenvalues of the FIM to uncover what they refer to as "pathological sharpness" or the distance between the mean value of the eigenvalues of the FIM and the maximum value of the FIM \cite{Karakida2_mse, Karakida3_crossentropy}. Seeking models with low sharpness in the loss landscape in the local neighborhood with respect to the parameters have been observed to improve generalization \cite{SAM}, so it is plausible that the FIM provides correlative information on model generalizability. In the layerwise setting, each operation of the neural network also can be used to create a [$p_l$ x $p_l$] layerwise Fisher information matrix. Given that we know $p_l$ from our architecture, once we calculate the layerwise NTK we actually know the entire spectra of these layerwise FIM. 


\subsection{Visualizing the NTK over training}
\label{VisNTKoverTraining}




In this experiment, we calculated the NTK and each layerwise NTK additive component for every training step of an MLP trained by vanilla gradient descent to classify MNIST-2, where we have randomly sampled handwritten digits of class 6 and 9. By collecting the NTK at every training step, we can reconstruct a video of how the NTK changes that you can view \href{https://youtu.be/Og796gMShcE}{ here.}. A more detailed explanation of our experimental setup are available in appendix \ref{ExperimentDetails}.

In the plots below, we have sorted our training data such that the first 5000 indices are all class 6 and the next 5000 indices are all class 9. This makes visualization more interpretable and does not impact learning because our gradient updates are averaged over the entire training dataset.

\begin{figure}[h]
\centering
\includegraphics[width=8cm, height=8cm]{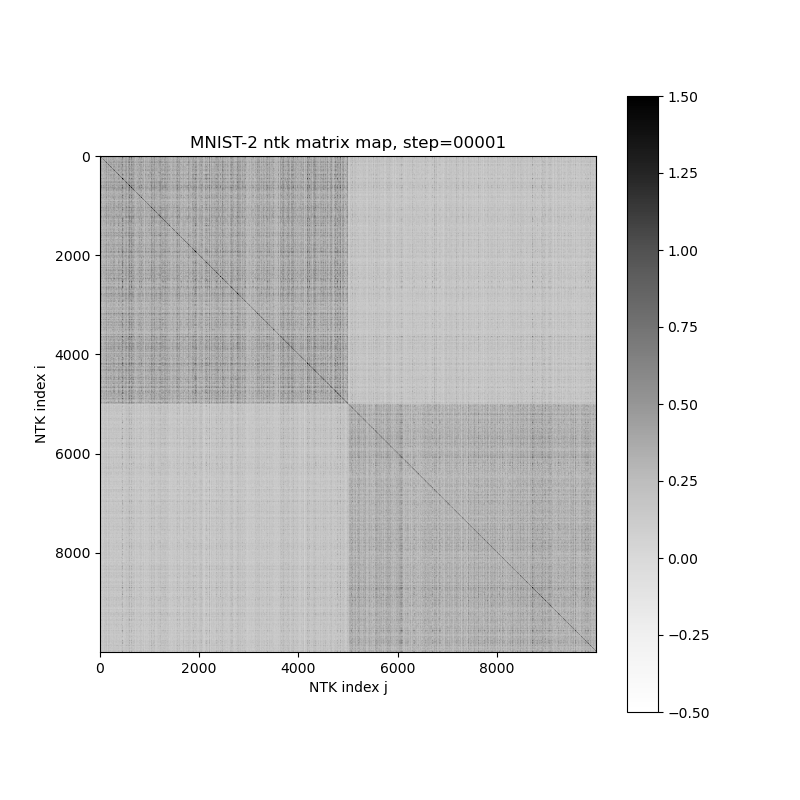}
\includegraphics[width=8cm,height=8cm]{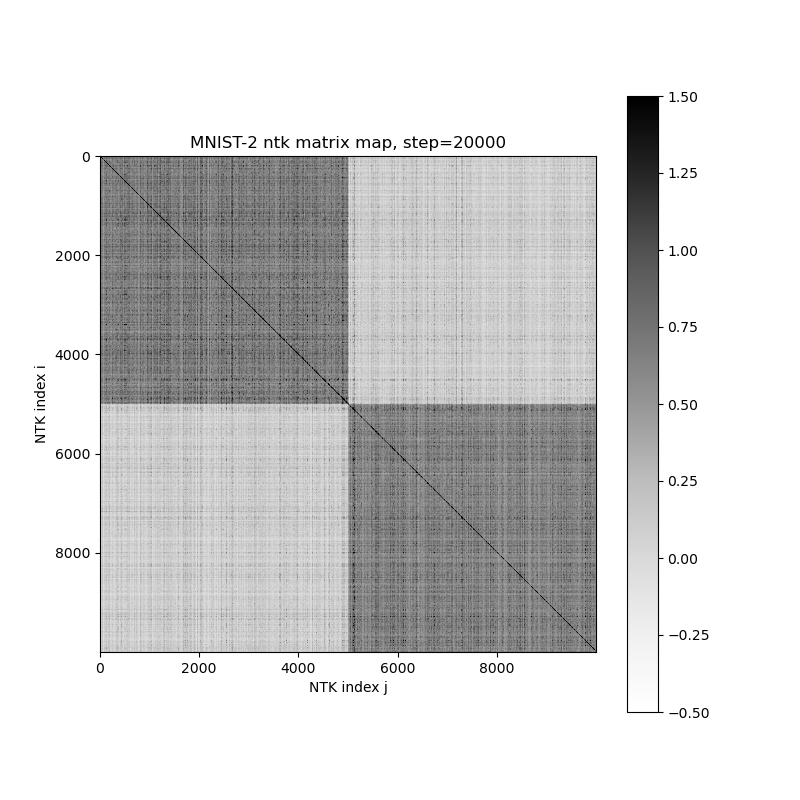}
\caption{The initial (left) and final (right) NTK matrix values; the final values were calculated after training the neural network for 20k epochs on a binary classification task. The dataset was sorted before training with all indices 0-5k from MNIST class 6 and 5k-10k from MNIST class 9. Therefore, the darker squares in the top left and bottom right quadrant are expected given similar datapoints are of the same class and that the NTK represents a similarity metric. This also explains the high values along the diagonal, where the NTK similarity between the same datapoints should be expected to be high. The full NTK shown in these images is constructed by summing the additive components from the parameterized operations of the network, plotted in the appendices as figures \ref{fig:MNIST2_l1} through \ref{fig:MNIST2_l4}}
\label{fig:MNIST2_ntk}
\end{figure}

In Fig \ref{fig:MNIST2_ntk} we plot the initial and final NTK matrix over training. The kernel has discriminatory ability, meaning that the block of 6s have in general higher NTK value than the block in the upper right and lower left quadrant. Note that as training progresses the diagonal blocks become darker and the off-diagonal blocks become lighter, representing that the NTK is capturing information about how the neural network is differentiating between classes. This is consistent with the intuition that NTK represents a similarity score between datapoints as measured by a dot product between the neural function's gradients. One can use this kernel to do binary classification by computing the similarity of some training point x with the dataset X. The kernel machine describing binary classification is:

$$ y_{i} = \mathrm{sgn}( \sum_{k} w_{k} Y_{k} K(x_{i},X_{k}) )$$

Where the result is mapped onto -1,1. Because our training data is balanced, and for simplicity, we set all $w_{k}$ to 1 as a quick approximation of the accuracy of a kernel machine that could utilize each NTK. We compute these accuracies at the start and end of training and report them in the table: \ref{T:layerwise_kernel_performance}

\begin{table}[!htbp]
\begin{center}
\begin{tabular}{| c | c | c |}
\hline
\textbf{Kernel/Method} & \textbf{initialization} & \textbf{training end}  \\ 
\cline{1-3}
layer 1 & 95.0 +/- 0.2 &  98.37 +/- 0.04 \\ \hline
layer 2 & 96.5 +/- 0.1  &  98.54 +/- 0.04 \\ \hline
layer 3 & 98.0 +/- 0.1  &  98.91 +/- 0.03 \\ \hline
layer 4 & 96.2 +/- 0.1  &  97.2 +/- 0.1 \\ \hline 
NTK & 98.56 +/- 0.04 &  98.96 +/- 0.03 \\ \hline \hline
Neural Network (train) & 51 +/- 1 & 98.81 +/- 0.02 \\ \hline
Neural Network (test) & 50 +/- 1 & 97.83 +/- 0.05 \\ \hline
\end{tabular}
\end{center}
\caption{The test accuracies from a holdout dataset from the simplified kernel machine are shown in the table before and after training. We note that all accuracies increased as the underlying network specialized to our training task, and that the final accuracy of the underlying ANN model exceeded the performance of any of the kernels. Another interesting observation is that not all layers have the same accuracy. While this is possibly explained between the layer 4 and layer 1 NTK as simply a matter of how many parameters are being used to measure the similarity, layer 2 and layer 3 have the exact same number of parameters, but still have a different performance. Future work will be conducted to systematically measure these performances from a variety of MLP architectures using this software package.}
\label{T:layerwise_kernel_performance}
\end{table}

While not theoretically precise, it is possible that quantities of the finite NTK can give insight to properties of the neural network, and in fact, there is preliminary evidence showing that these finite kernels (non-linearly) correlate with the performance of their infinite width counterparts in CNNs, which in turn themselves correlate with the ANN's performance (compare table 2 and table 1 of \cite{CNTK_Sanjeev}).  Using this fact, one might be able to initiate a neural architecture search by searching for architectures or parameterizations whose initial NTK gives better performance.

\section{Future Work}

\subsection{Future Improvements and Known Issues}

We are releasing our software in alpha open-source with a pledge to continue to improve and update our software. We welcome the contributions of the community and look forward to see how other groups might use or be inspired by the software. There are specific improvements to make the software complete that we briefly touch on in this section.

Currently, each algorithm expects a single neuron output. This is a significant limitation, as common practice for even basic multi-task classification would be to have a number of output neurons representing each class. We believe that our autograd techniques could be extended to multiple output neurons with additional effort.

Motivated by our explicit derivative success in MLP, we could add additional derivatives for other architectures. Initial attempts at extending an explicit derivative to fully convolutional networks became memory inefficient by relying on large matrices to describe derivatives of the convolution operation. However, additional effort could be placed towards the end of achieving a fast and memory efficient form.

The software lacks multi-GPU support. Note that the neural-tangents library includes native GPU parallelization. Multiple GPU support would be a large boon; it targets two core issues with NTKs for larger models-- memory constraints and time costs. Even the calculation of the NTK for a modest multilayer perceptron on a subset of MNIST requires the full memory of a single A100 GPU (see Table \ref{T:MLP Benchmark Memory}). This means for more common workstations and consumer level GPUs researchers are still severely limited to small models and small datasets.

\section{Conclusion}

This technical report has described the theoretical background and functional performance of torchNTK. This software has the capability to efficiently calculate the tangent kernel for MLP architectures in PyTorch, but through autograd methods we have extended the utility to arbitrary architectures. This work is impactful because neural kernels are objects of interest to the theoretical community, and with PyTorch support we can extend the number of researchers who have access to compute them. Our software enables teams to more easily calculate the kernels, which we hope will give way for further research and application.

A key takeaway from our work is that teams should consider the performance needs to conduct their research and determine whether calculating the explicit derivative of the network with respect to the parameters is worthwhile. We have shown that, at-least in the cases of MLP architectures, explicit differentiation is more efficient in both time and memory than autograd methods. Furthermore; teams must evaluate honestly whether they have access to the software expertise to implement the calculation they derive in the most parallelized or efficient manner. Converting the derived equations to efficient code is a skill set that should not be underestimated.

This software enables researchers to calculate the NTK in PyTorch faster than ever before and exposes the user to what we have called the layerwise components of the NTK, each representing a parameterized operation inside the neural network. Our future work will wield this software package to explore the NTK and these components to search for use-cases for practical A.I. end users and interpretation of such models.

\printbibliography

@ARTICLE{Zhichaos_paper,
       author = {{Fan}, Zhou and {Wang}, Zhichao},
        title = "{Spectra of the Conjugate Kernel and Neural Tangent Kernel for linear-width neural networks}",
      journal = {arXiv e-prints},
         year = 2020,
        month = may,
          eid = {arXiv:2005.11879},
        pages = {arXiv:2005.11879},
archivePrefix = {arXiv},
       eprint = {2005.11879},
 primaryClass = {stat.ML},
       adsurl = {https://ui.adsabs.harvard.edu/abs/2020arXiv200511879F}
}

@ARTICLE{TENAS,
       author = {{Chen}, Wuyang and {Gong}, Xinyu and {Wang}, Zhangyang},
        title = "{Neural Architecture Search on ImageNet in Four GPU Hours: A Theoretically Inspired Perspective}",
      journal = {arXiv e-prints},
         year = 2021,
        month = feb,
          eid = {arXiv:2102.11535},
        pages = {arXiv:2102.11535},
archivePrefix = {arXiv},
       eprint = {2102.11535},
 primaryClass = {cs.CV},
       adsurl = {https://ui.adsabs.harvard.edu/abs/2021arXiv210211535C}
}

@ARTICLE{Domingos21,
       author = {{Domingos}, Pedro},
        title = "{Every Model Learned by Gradient Descent Is Approximately a Kernel Machine}",
      journal = {arXiv e-prints},
         year = 2020,
        month = nov,
          eid = {arXiv:2012.00152},
        pages = {arXiv:2012.00152},
archivePrefix = {arXiv},
       eprint = {2012.00152},
 primaryClass = {cs.LG},
       adsurl = {https://ui.adsabs.harvard.edu/abs/2020arXiv201200152D}
}

@ARTICLE{neural_tangents,
       author = {{Novak}, Roman and {Xiao}, Lechao and {Hron}, Jiri and {Lee}, Jaehoon and {Alemi}, Alexander A. and {Sohl-Dickstein}, Jascha and {Schoenholz}, Samuel S.},
        title = "{Neural Tangents: Fast and Easy Infinite Neural Networks in Python}",
      journal = {arXiv e-prints},
         year = 2019,
        month = dec,
          eid = {arXiv:1912.02803},
        pages = {arXiv:1912.02803},
archivePrefix = {arXiv},
       eprint = {1912.02803},
 primaryClass = {stat.ML},
       adsurl = {https://ui.adsabs.harvard.edu/abs/2019arXiv191202803N}
}

@incollection{PyTorch,
title = {PyTorch: An Imperative Style, High-Performance Deep Learning Library},
author = {Paszke, Adam and Gross, Sam and Massa, Francisco and Lerer, Adam and Bradbury, James and Chanan, Gregory and Killeen, Trevor and Lin, Zeming and Gimelshein, Natalia and Antiga, Luca and Desmaison, Alban and Kopf, Andreas and Yang, Edward and DeVito, Zachary and Raison, Martin and Tejani, Alykhan and Chilamkurthy, Sasank and Steiner, Benoit and Fang, Lu and Bai, Junjie and Chintala, Soumith},
booktitle = {Advances in Neural Information Processing Systems 32},
editor = {H. Wallach and H. Larochelle and A. Beygelzimer and F. d\textquotesingle Alche-Buc and E. Fox and R. Garnett},
pages = {8024--8035},
year = {2019},
publisher = {Curran Associates, Inc.},
url = {http://papers.neurips.cc/paper/9015-pytorch-an-imperative-style-high-performance-deep-learning-library.pdf}
}

@ARTICLE{resnet2015,
       author = {{He}, Kaiming and {Zhang}, Xiangyu and {Ren}, Shaoqing and {Sun}, Jian},
        title = "{Deep Residual Learning for Image Recognition}",
      journal = {arXiv e-prints},
         year = 2015,
        month = dec,
          eid = {arXiv:1512.03385},
        pages = {arXiv:1512.03385},
archivePrefix = {arXiv},
       eprint = {1512.03385},
 primaryClass = {cs.CV},
       adsurl = {https://ui.adsabs.harvard.edu/abs/2015arXiv151203385H}
}

@ARTICLE{GPT3,
       author = {{Brown}, Tom B. and {Mann}, Benjamin and {Ryder}, Nick and {Subbiah}, Melanie and {Kaplan}, Jared and {Dhariwal}, Prafulla and {Neelakantan}, Arvind and {Shyam}, Pranav and {Sastry}, Girish and {Askell}, Amanda and {Agarwal}, Sandhini and {Herbert-Voss}, Ariel and {Krueger}, Gretchen and {Henighan}, Tom and {Child}, Rewon and {Ramesh}, Aditya and {Ziegler}, Daniel M. and {Wu}, Jeffrey and {Winter}, Clemens and {Hesse}, Christopher and {Chen}, Mark and {Sigler}, Eric and {Litwin}, Mateusz and {Gray}, Scott and {Chess}, Benjamin and {Clark}, Jack and {Berner}, Christopher and {McCandlish}, Sam and {Radford}, Alec and {Sutskever}, Ilya and {Amodei}, Dario},
        title = "{Language Models are Few-Shot Learners}",
      journal = {arXiv e-prints},
         year = 2020,
        month = may,
          eid = {arXiv:2005.14165},
        pages = {arXiv:2005.14165},
archivePrefix = {arXiv},
       eprint = {2005.14165},
 primaryClass = {cs.CL},
       adsurl = {https://ui.adsabs.harvard.edu/abs/2020arXiv200514165B}
}

@software{jaxgithub,
  author = {James Bradbury and Roy Frostig and Peter Hawkins and Matthew James Johnson and Chris Leary and Dougal Maclaurin and George Necula and Adam Paszke and Jake Vander{P}las and Skye Wanderman-{M}ilne and Qiao Zhang},
  title = {{JAX}: composable transformations of {P}ython+{N}um{P}y programs},
  url = {http://github.com/google/jax},
  version = {0.2.5},
  year = {2018},
}

@misc{kaggle_2021,
 title={State of Data Science and Machine Learning 2021},
 url={https://www.kaggle.com/kaggle-survey-2021},
 journal={Kaggle},
 year={2021},
 month={10}
 }

@ARTICLE{Jacot_NTK2018_Convergence,
       author = {{Jacot}, Arthur and {Gabriel}, Franck and {Hongler}, Clement},
        title = "{Neural Tangent Kernel: Convergence and Generalization in Neural Networks}",
      journal = {arXiv e-prints},
         year = 2018,
        month = jun,
          eid = {arXiv:1806.07572},
        pages = {arXiv:1806.07572},
archivePrefix = {arXiv},
       eprint = {1806.07572},
 primaryClass = {cs.LG},
       adsurl = {https://ui.adsabs.harvard.edu/abs/2018arXiv180607572J}
}

@ARTICLE{CNTK_Sanjeev,
       author = {{Arora}, Sanjeev and {Du}, Simon S. and {Hu}, Wei and {Li}, Zhiyuan and {Salakhutdinov}, Ruslan and {Wang}, Ruosong},
        title = "{On Exact Computation with an Infinitely Wide Neural Net}",
      journal = {arXiv e-prints},
         year = 2019,
        month = apr,
          eid = {arXiv:1904.11955},
        pages = {arXiv:1904.11955},
archivePrefix = {arXiv},
       eprint = {1904.11955},
 primaryClass = {cs.LG},
       adsurl = {https://ui.adsabs.harvard.edu/abs/2019arXiv190411955A}
}

@InProceedings{TensorPrograms4,
  title = 	 {Tensor Programs IV: Feature Learning in Infinite-Width Neural Networks},
  author =       {Yang, Greg and Hu, Edward J.},
  booktitle = 	 {Proceedings of the 38th International Conference on Machine Learning},
  pages = 	 {11727--11737},
  year = 	 {2021},
  editor = 	 {Meila, Marina and Zhang, Tong},
  volume = 	 {139},
  series = 	 {Proceedings of Machine Learning Research},
  month = 	 {18--24 Jul},
  publisher =    {PMLR},
  url = 	 {https://proceedings.mlr.press/v139/yang21c.html}
}

@inproceedings{AlexNet2012,
 author = {Krizhevsky, Alex and Sutskever, Ilya and Hinton, Geoffrey E},
 booktitle = {Advances in Neural Information Processing Systems},
 editor = {F. Pereira and C. J. C. Burges and L. Bottou and K. Q. Weinberger},
 pages = {},
 publisher = {Curran Associates, Inc.},
 title = {ImageNet Classification with Deep Convolutional Neural Networks},
 url = {https://proceedings.neurips.cc/paper/2012/file/c399862d3b9d6b76c8436e924a68c45b-Paper.pdf},
 volume = {25},
 year = {2012}
}

@ARTICLE{ViTOG,
       author = {{Dosovitskiy}, Alexey and {Beyer}, Lucas and {Kolesnikov}, Alexander and {Weissenborn}, Dirk and {Zhai}, Xiaohua and {Unterthiner}, Thomas and {Dehghani}, Mostafa and {Minderer}, Matthias and {Heigold}, Georg and {Gelly}, Sylvain and {Uszkoreit}, Jakob and {Houlsby}, Neil},
        title = "{An Image is Worth 16x16 Words: Transformers for Image Recognition at Scale}",
      journal = {arXiv e-prints},
         year = 2020,
        month = oct,
          eid = {arXiv:2010.11929},
        pages = {arXiv:2010.11929},
archivePrefix = {arXiv},
       eprint = {2010.11929},
 primaryClass = {cs.CV},
       adsurl = {https://ui.adsabs.harvard.edu/abs/2020arXiv201011929D}
}

@ARTICLE{EffNetV2OG,
       author = {{Tan}, Mingxing and {Le}, Quoc V.},
        title = "{EfficientNetV2: Smaller Models and Faster Training}",
      journal = {arXiv e-prints},
         year = 2021,
        month = apr,
          eid = {arXiv:2104.00298},
        pages = {arXiv:2104.00298},
archivePrefix = {arXiv},
       eprint = {2104.00298},
 primaryClass = {cs.CV},
       adsurl = {https://ui.adsabs.harvard.edu/abs/2021arXiv210400298T}
}

@ARTICLE{InceptionResNetV4,
       author = {{Szegedy}, Christian and {Ioffe}, Sergey and {Vanhoucke}, Vincent and {Alemi}, Alex},
        title = "{Inception-v4, Inception-ResNet and the Impact of Residual Connections on Learning}",
      journal = {arXiv e-prints},
         year = 2016,
        month = feb,
          eid = {arXiv:1602.07261},
        pages = {arXiv:1602.07261},
archivePrefix = {arXiv},
       eprint = {1602.07261},
 primaryClass = {cs.CV},
       adsurl = {https://ui.adsabs.harvard.edu/abs/2016arXiv160207261S}
}

@inproceedings{cupy,
  author = {{Okuta}, Ryosuke and {Unno}, Yuya and {Nishino}, Daisuke and {Hido}, Shohei and {Loomis}, Crissman},
  title        = {CuPy: A NumPy-Compatible Library for NVIDIA GPU Calculations},
  booktitle    = {Proceedings of Workshop on Machine Learning Systems (LearningSys) in The Thirty-first Annual Conference on Neural Information Processing Systems (NIPS)},
  year         = {2017},
  url          = {http://learningsys.org/nips17/assets/papers/paper_16.pdf}
}

@ARTICLE{LayerwiseHessian,
       author = {{Sankar}, Adepu Ravi and {Khasbage}, Yash and {Vigneswaran}, Rahul and {Balasubramanian}, Vineeth N},
        title = "{A Deeper Look at the Hessian Eigenspectrum of Deep Neural Networks and its Applications to Regularization}",
      journal = {arXiv e-prints},
         year = 2020,
        month = dec,
          eid = {arXiv:2012.03801},
        pages = {arXiv:2012.03801},
archivePrefix = {arXiv},
       eprint = {2012.03801},
 primaryClass = {cs.LG},
       adsurl = {https://ui.adsabs.harvard.edu/abs/2020arXiv201203801S}
}

@misc{tensorflow,
title={ {TensorFlow}: Large-Scale Machine Learning on Heterogeneous Systems},
url={https://www.tensorflow.org/},
note={Software available from tensorflow.org},
author={
    Martin~Abadi and
    Ashish~Agarwal and
    Paul~Barham and
    Eugene~Brevdo and
    Zhifeng~Chen and
    Craig~Citro and
    Greg~S.~Corrado and
    Andy~Davis and
    Jeffrey~Dean and
    Matthieu~Devin and
    Sanjay~Ghemawat and
    Ian~Goodfellow and
    Andrew~Harp and
    Geoffrey~Irving and
    Michael~Isard and
    Yangqing Jia and
    Rafal~Jozefowicz and
    Lukasz~Kaiser and
    Manjunath~Kudlur and
    Josh~Levenberg and
    Dandelion~Mane and
    Rajat~Monga and
    Sherry~Moore and
    Derek~Murray and
    Chris~Olah and
    Mike~Schuster and
    Jonathon~Shlens and
    Benoit~Steiner and
    Ilya~Sutskever and
    Kunal~Talwar and
    Paul~Tucker and
    Vincent~Vanhoucke and
    Vijay~Vasudevan and
    Fernanda~Viegas and
    Oriol~Vinyals and
    Pete~Warden and
    Martin~Wattenberg and
    Martin~Wicke and
    Yuan~Yu and
    Xiaoqiang~Zheng},
  year={2015},
}

@ARTICLE{Karakida2_mse,
       author = {{Karakida}, Ryo and {Akaho}, Shotaro and {Amari}, Shun-ichi},
        title = "{The Normalization Method for Alleviating Pathological Sharpness in Wide Neural Networks}",
      journal = {arXiv e-prints},
         year = 2019,
        month = jun,
          eid = {arXiv:1906.02926},
        pages = {arXiv:1906.02926},
archivePrefix = {arXiv},
       eprint = {1906.02926},
 primaryClass = {stat.ML},
       adsurl = {https://ui.adsabs.harvard.edu/abs/2019arXiv190602926K}
}

@ARTICLE{Karakida3_crossentropy,
       author = {{Karakida}, Ryo and {Akaho}, Shotaro and {Amari}, Shun-ichi},
        title = "{Pathological spectra of the Fisher information metric and its variants in deep neural networks}",
      journal = {arXiv e-prints},
         year = 2019,
        month = oct,
          eid = {arXiv:1910.05992},
        pages = {arXiv:1910.05992},
archivePrefix = {arXiv},
       eprint = {1910.05992},
 primaryClass = {stat.ML},
       adsurl = {https://ui.adsabs.harvard.edu/abs/2019arXiv191005992K}
}

@ARTICLE{Lee2019_empricalNTK,
       author = {{Lee}, Jaehoon and {Xiao}, Lechao and {Schoenholz}, Samuel S. and {Bahri}, Yasaman and {Novak}, Roman and {Sohl-Dickstein}, Jascha and {Pennington}, Jeffrey},
        title = "{Wide neural networks of any depth evolve as linear models under gradient descent}",
      journal = {Journal of Statistical Mechanics: Theory and Experiment},
         year = 2020,
        month = dec,
       volume = {2020},
       number = {12},
          eid = {124002},
        pages = {124002},
          doi = {10.1088/1742-5468/abc62b},
archivePrefix = {arXiv},
       eprint = {1902.06720},
 primaryClass = {stat.ML},
       adsurl = {https://ui.adsabs.harvard.edu/abs/2020JSMTE2020l4002L}
}

@ARTICLE{MartinMahoneyNature,
       author = {{Martin}, Charles H. and {Peng}, Tongsu Serena and {Mahoney}, Michael W.},
        title = "{Predicting trends in the quality of state-of-the-art neural networks without access to training or testing data}",
      journal = {Nature Communications},
         year = 2021,
        month = jan,
       volume = {12},
          eid = {4122},
        pages = {4122},
          doi = {10.1038/s41467-021-24025-8},
archivePrefix = {arXiv},
       eprint = {2002.06716},
 primaryClass = {cs.LG},
       adsurl = {https://ui.adsabs.harvard.edu/abs/2021NatCo..12.4122M}
}

@ARTICLE{Layerwise_Hessian,
       author = {{Wu}, Yikai and {Zhu}, Xingyu and {Wu}, Chenwei and {Wang}, Annie and {Ge}, Rong},
        title = "{Dissecting Hessian: Understanding Common Structure of Hessian in Neural Networks}",
      journal = {arXiv e-prints},
         year = 2020,
        month = oct,
          eid = {arXiv:2010.04261},
        pages = {arXiv:2010.04261},
archivePrefix = {arXiv},
       eprint = {2010.04261},
 primaryClass = {cs.LG},
       adsurl = {https://ui.adsabs.harvard.edu/abs/2020arXiv201004261W}
}

@ARTICLE{SAM,
       author = {{Foret}, Pierre and {Kleiner}, Ariel and {Mobahi}, Hossein and {Neyshabur}, Behnam},
        title = "{Sharpness-Aware Minimization for Efficiently Improving Generalization}",
      journal = {arXiv e-prints},
         year = 2020,
        month = oct,
          eid = {arXiv:2010.01412},
        pages = {arXiv:2010.01412},
archivePrefix = {arXiv},
       eprint = {2010.01412},
 primaryClass = {cs.LG},
       adsurl = {https://ui.adsabs.harvard.edu/abs/2020arXiv201001412F}
}

@ARTICLE{FinitevsEmpirical2020,
       author = {{Lee}, Jaehoon and {Schoenholz}, Samuel S. and {Pennington}, Jeffrey and {Adlam}, Ben and {Xiao}, Lechao and {Novak}, Roman and {Sohl-Dickstein}, Jascha},
        title = "{Finite Versus Infinite Neural Networks: an Empirical Study}",
      journal = {arXiv e-prints},
         year = 2020,
        month = jul,
          eid = {arXiv:2007.15801},
        pages = {arXiv:2007.15801},
archivePrefix = {arXiv},
       eprint = {2007.15801},
 primaryClass = {cs.LG},
       adsurl = {https://ui.adsabs.harvard.edu/abs/2020arXiv200715801L}
}

@ARTICLE{HarnessingThePowerArora2020,
       author = {{Arora}, Sanjeev and {Du}, Simon S. and {Li}, Zhiyuan and {Salakhutdinov}, Ruslan and {Wang}, Ruosong and {Yu}, Dingli},
        title = "{Harnessing the Power of Infinitely Wide Deep Nets on Small-data Tasks}",
      journal = {arXiv e-prints},
         year = 2019,
        month = oct,
          eid = {arXiv:1910.01663},
        pages = {arXiv:1910.01663},
archivePrefix = {arXiv},
       eprint = {1910.01663},
 primaryClass = {cs.LG},
       adsurl = {https://ui.adsabs.harvard.edu/abs/2019arXiv191001663A}
}

@ARTICLE{LazyTraining,
       author = {{Chizat}, Lenaic and {Oyallon}, Edouard and {Bach}, Francis},
        title = "{On Lazy Training in Differentiable Programming}",
      journal = {arXiv e-prints},
         year = 2018,
        month = dec,
          eid = {arXiv:1812.07956},
        pages = {arXiv:1812.07956},
archivePrefix = {arXiv},
       eprint = {1812.07956},
 primaryClass = {math.OC},
       adsurl = {https://ui.adsabs.harvard.edu/abs/2018arXiv181207956C}
}

@ARTICLE{DisentanglingFeatureLazy,
       author = {{Geiger}, Mario and {Spigler}, Stefano and {Jacot}, Arthur and {Wyart}, Matthieu},
        title = "{Disentangling feature and lazy training in deep neural networks}",
      journal = {Journal of Statistical Mechanics: Theory and Experiment},
         year = 2020,
        month = nov,
       volume = {2020},
       number = {11},
          eid = {113301},
        pages = {113301},
          doi = {10.1088/1742-5468/abc4de},
archivePrefix = {arXiv},
       eprint = {1906.08034},
 primaryClass = {cs.LG},
       adsurl = {https://ui.adsabs.harvard.edu/abs/2020JSMTE2020k3301G}
}

@ARTICLE{WideNNevolveaslinear,
       author = {{Lee}, Jaehoon and {Xiao}, Lechao and {Schoenholz}, Samuel S. and {Bahri}, Yasaman and {Novak}, Roman and {Sohl-Dickstein}, Jascha and {Pennington}, Jeffrey},
        title = "{Wide neural networks of any depth evolve as linear models under gradient descent}",
      journal = {Journal of Statistical Mechanics: Theory and Experiment},
         year = 2020,
        month = dec,
       volume = {2020},
       number = {12},
          eid = {124002},
        pages = {124002},
          doi = {10.1088/1742-5468/abc62b},
archivePrefix = {arXiv},
       eprint = {1902.06720},
 primaryClass = {stat.ML},
       adsurl = {https://ui.adsabs.harvard.edu/abs/2020JSMTE2020l4002L}
}

@ARTICLE{LayerwiseSpectral,
       author = {{Dandi}, Yatin and {Jacot}, Arthur},
        title = "{Understanding Layer-wise Contributions in Deep Neural Networks through Spectral Analysis}",
      journal = {arXiv e-prints},
         year = 2021,
        month = nov,
          eid = {arXiv:2111.03972},
        pages = {arXiv:2111.03972},
archivePrefix = {arXiv},
       eprint = {2111.03972},
 primaryClass = {cs.LG},
       adsurl = {https://ui.adsabs.harvard.edu/abs/2021arXiv211103972D}
}

\pagebreak

\appendix

\section{Layerwise NTK visualizations}
Plotted below are the additive components of the NTK for the experiment described in Section \ref{VisNTKoverTraining}. Figure \ref{fig:MNIST2_ntk} described the evolution of the full NTK over training, here we describe the evolution of the additive components of the NTK for the same experiment.

\begin{figure}[!htb]
\centering
\includegraphics[width=8cm, height=8cm]{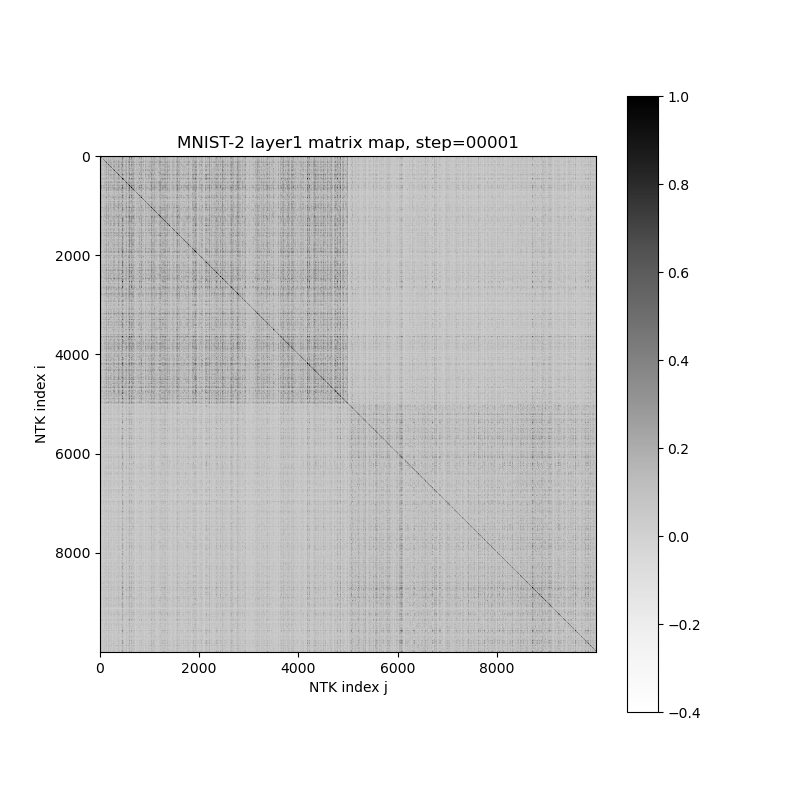}
\includegraphics[width=8cm,height=8cm]{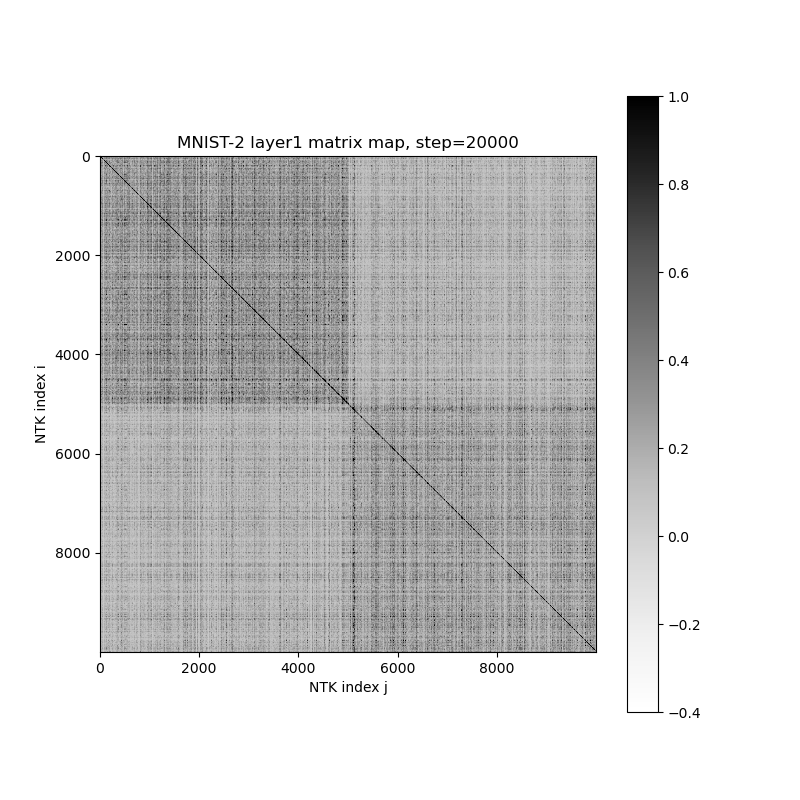}
\caption{initial (left) and final (right) layerwise NTK for layer 1, the first dense layer. This layer had 39200 parameters. While the contrast increased between the datapoints of different classes similar to the remaining layers, layer 1 is unique in that the location of horizontal and vertical bar features shifts over training. We hypothesize this is a shift in what the NTK measures as an outlier}
\label{fig:MNIST2_l1}
\end{figure}

\begin{figure}[!htb]
\centering
\includegraphics[width=8cm, height=8cm]{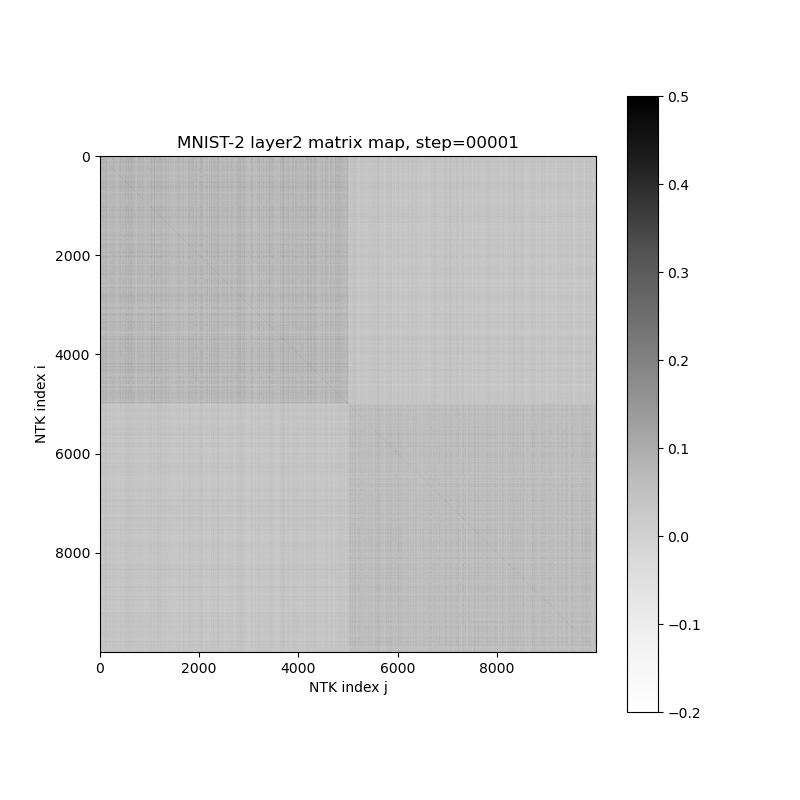}
\includegraphics[width=8cm,height=8cm]{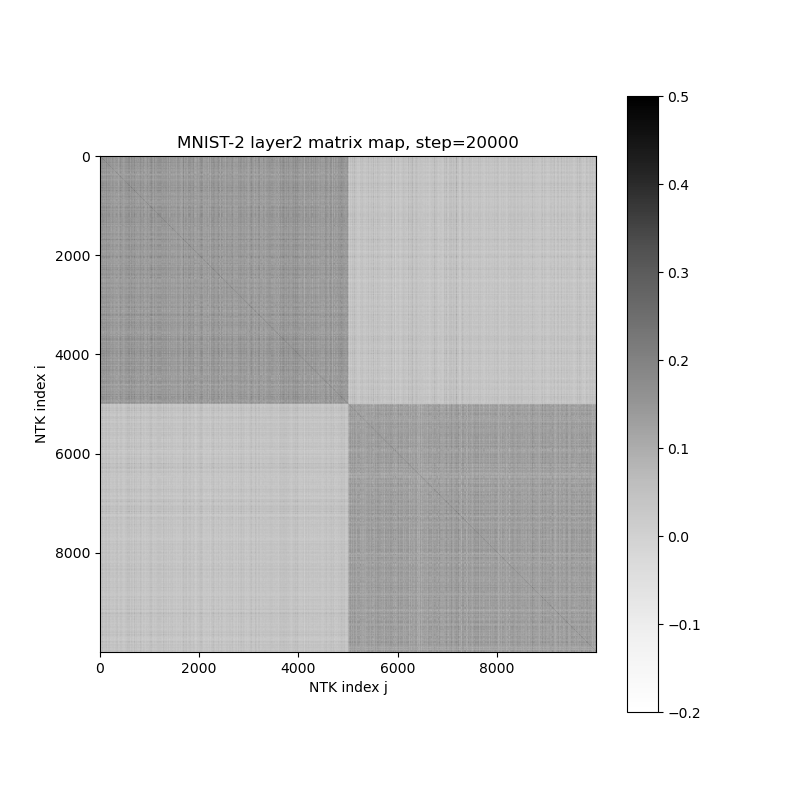}
\caption{initial (left) and final (right) layerwise NTK for layer 2, the second dense layer. Layer 2 has 2500 parameters. Generally, the notable behavior over the course of training was that in layer 2 the contrast increases between examples of MNIST class 6s and MNIST class 9s.}
\label{fig:MNIST2_l2}
\end{figure}

\begin{figure}[!htb]
\centering
\includegraphics[width=8cm, height=8cm]{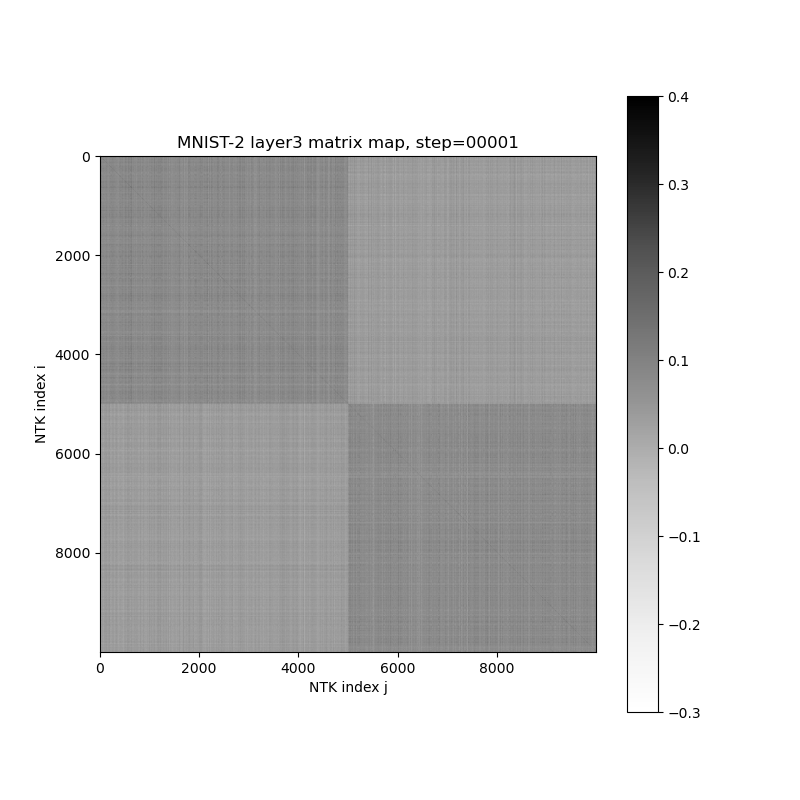}
\includegraphics[width=8cm,height=8cm]{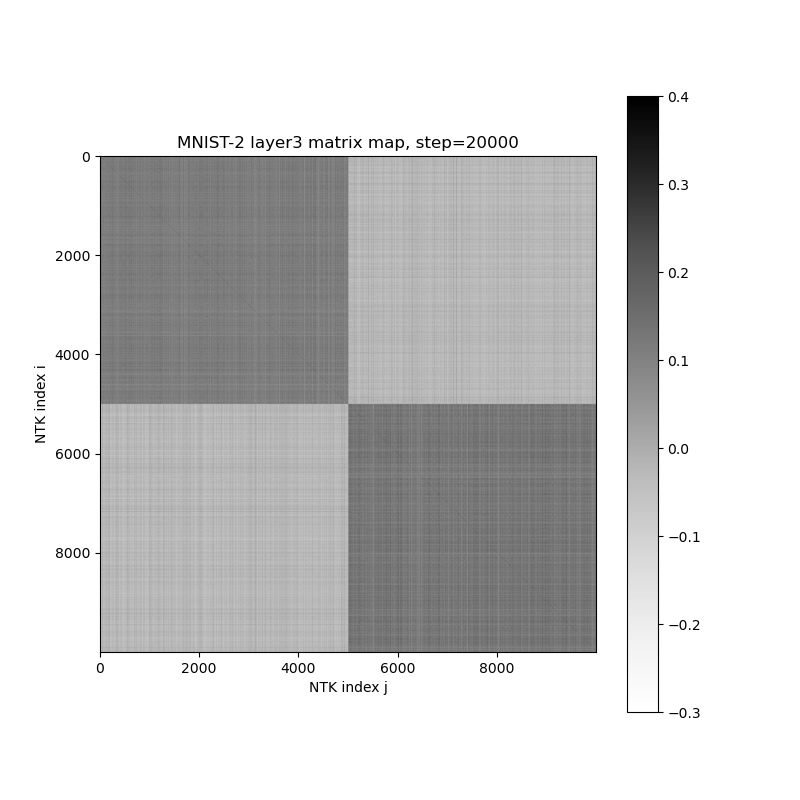}
\caption{initial (left) and final (right) layerwise NTK for layer 3, the third dense layer. Layer 3 has 2500 parameters. Generally, the notable behavior over the course of training was that in layer 3 the contrast increases between examples of MNIST class 6s and MNIST class 9s.}
\label{fig:MNIST2_l3}
\end{figure}

\begin{figure}[!h]
\centering
\includegraphics[width=8cm, height=8cm]{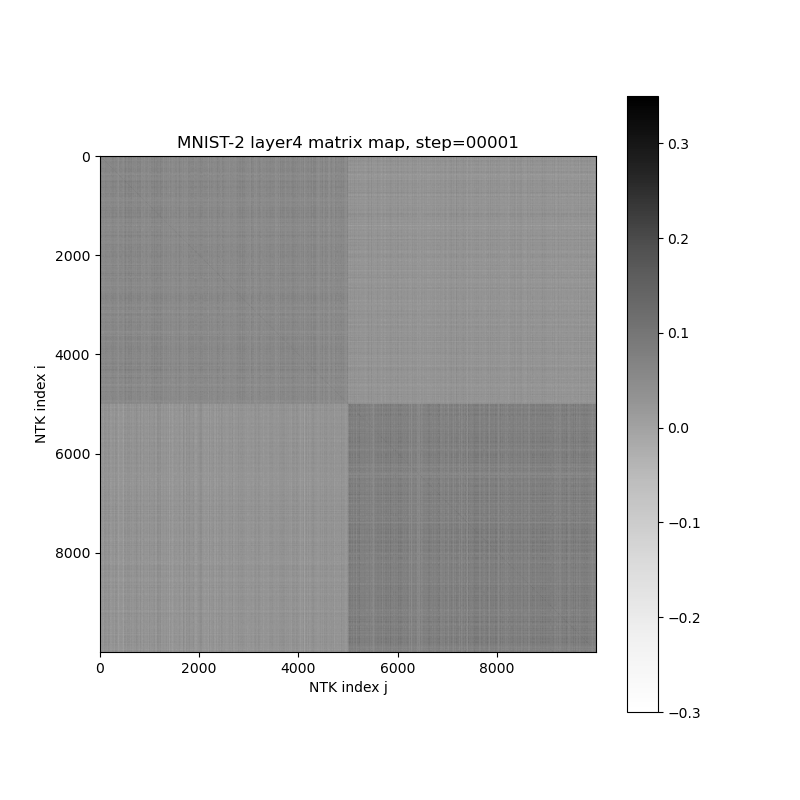}
\includegraphics[width=8cm,height=8cm]{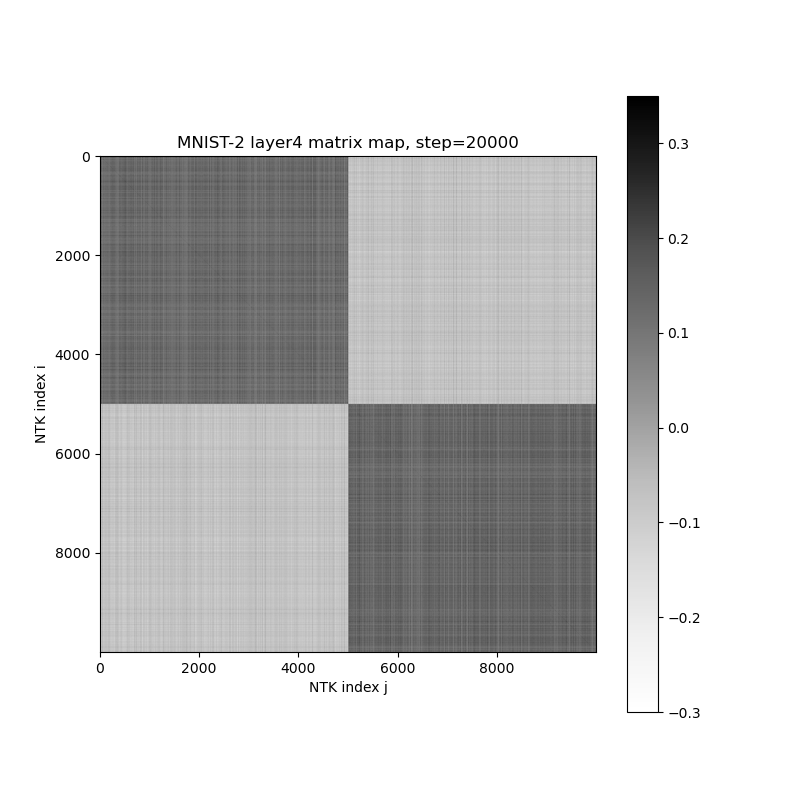}
\caption{initial (left) and final (right) layerwise NTK for layer 4, the fourth and final dense layer. Layer 4 has 50 parameters. Generally, the notable behavior over the course of training was that in layer 4 the contrast increases between examples of MNIST class 6s and MNIST class 9s.}
\label{fig:MNIST2_l4}
\end{figure}

\pagebreak
 \section{Details of Experiment in S\ref{VisNTKoverTraining}}
 \label{ExperimentDetails}
 
Our model is a four layer MLP with an input feature vector of size 784, each hidden layer has a width of 50 neurons, and ends in a single neuron readout layer to facilitate binary classification. The NTK is calculated before the final sigmoid activation, using explicit differentiation. Sigmoid was chosen to map the network function onto a binary decision between classes. The network was placed into the NTK parameterization by dividing each hidden layer by the square root of the width of the layer. Weights were initialized from the standard normal distribution, but biases were frozen at zero and not computed in the NTK. 

Our training dataset of 5000 examples of MNIST 6s and and 5000 9s were flattened to a feature vector, placed into the range 0-1, and normalized. The training dataset was sorted such that the first 5000 indices were all label 6 and the last 5000 were all label 9. Since we are using full-batch gradient descent sorting doesn't affect training, but makes visualizing the NTK easier. The model was trained for 20,000 gradient descent steps with a learning rate 1e-2.  At this point training had saturated but had not converged to a training loss of 0. Because we wanted to capture the NTK at every single update step, we believed it was prudent to stop training early at that point.

\pagebreak
\section{Example Usage}
Checkout the notebooks provided in the repository, especially, "DemoMethods.ipynb" for an overview of how to set the inputs for each individual algorithm; the notebook also demonstrates the agreement between methods for a small example. Below, we include a snippet that demonstrates the simplest and most general calculation of the layerwise NTK.

\begin{lstlisting}
#Example Use of autograd_components_ntk
#NOTE! should work for arbitrary architectures so long
#As the architecture terminates to a single neuron

from torchntk.autograd import autograd_components_ntk

class FC(torch.nn.Module):
    '''
    simple network
    '''
    def __init__(self,):
        super(FC, self).__init__()
        self.d1 = torch.nn.Linear(100,100,bias=False)
        self.d2 = torch.nn.Linear(100,100,bias=False)
        self.d3 = torch.nn.Linear(100,1,bias=False)
        
    def forward(self, x0):
        x1 = 1/np.sqrt(100) * activation(self.d1(x0))
        x2 = 1/np.sqrt(100) * activation(self.d2(x1))
        x3 =  self.d3(x2)
        return x3
        
model = FC()

train_x = torch.empty((200,100),device='cpu').normal_(0,1)
#train_x has shape [batch_size, n_features]

y = model(train_x.to(device))

#NTK_components is a dictionary with keys that are named
#parameters from the model and values that are NTK 
#created from those parameters' gradients.

NTK_components = autograd_components_ntk(model,y[:,0])

#Therefore, to get the full NTK you simply sum over
#the value tensors in the dictionary
NTK_full = torch.sum(torch.stack(
    [val for val in NTK_components.values()]
    ),dim=0)

\end{lstlisting}

\end{document}